\documentclass[10pt,twocolumn,letterpaper]{article}

\usepackage[pagenumbers]{cvpr} 

\usepackage{multirow}
\usepackage{booktabs}
\usepackage{graphicx}
\usepackage{amssymb} 
\usepackage{dsfont} 
\usepackage{url} 

\definecolor{cvprblue}{rgb}{0.21,0.49,0.74}
\usepackage[pagebackref,breaklinks,colorlinks,allcolors=cvprblue]{hyperref}

\def\paperID{*****} 
\def\confName{CVPR}
\def\confYear{2026}

\title{When Identities Collapse: A Stress-Test Benchmark for Multi-Subject Personalization}

\author{Zhihan Chen\thanks{Equal contribution}\\
University of California, Los Angeles\\
{\tt\small chenz23@ucla.edu}
\and
Yuhuan Zhao\footnotemark[1]\\
University of Southern California\\
{\tt\small yuhuanzh@usc.edu}\\
\and
Yijie Zhu\footnotemark[1]\\
DeerLab LLC\\
{\tt\small ejzhu2025@gmail.com}
\and
Xinyu Yao\footnotemark[1]\\
Carnegie Mellon University\\
{\tt\small xinyuyao@andrew.cmu.edu}
}
\begin{document}
\maketitle
\begin{abstract}
Subject-driven text-to-image diffusion models have achieved remarkable success in preserving single identities, yet their ability to compose multiple interacting subjects remains largely unexplored and highly challenging. Existing evaluation protocols typically rely on global CLIP metrics, which are insensitive to local identity collapse and fail to capture the severity of multi-subject entanglement. In this paper, we identify a pervasive ``Illusion of Scalability'' in current models: while they excel at synthesizing 2-4 subjects in simple layouts, they suffer from catastrophic identity collapse when scaled to 6-10 subjects or tasked with complex physical interactions. To systematically expose this failure mode, we construct a rigorous stress-test benchmark comprising 75 prompts distributed across varying subject counts and interaction difficulties (Neutral, Occlusion, Interaction). Furthermore, we demonstrate that standard CLIP-based metrics are fundamentally flawed for this task, as they often assign high scores to semantically correct but identity-collapsed images (e.g., generating generic clones). To address this, we introduce the Subject Collapse Rate (SCR), a novel evaluation metric grounded in DINOv2's structural priors, which strictly penalizes local attention leakage and homogenization. Our extensive evaluation of state-of-the-art models (MOSAIC, XVerse, PSR) reveals a precipitous drop in identity fidelity as scene complexity grows, with SCR approaching 100\% at 10 subjects. We trace this collapse to the semantic shortcuts inherent in global attention routing, underscoring the urgent need for explicit physical disentanglement in future generative architectures.
\end{abstract}

\begin{figure*}[t]
  \centering
  \includegraphics[width=\textwidth]{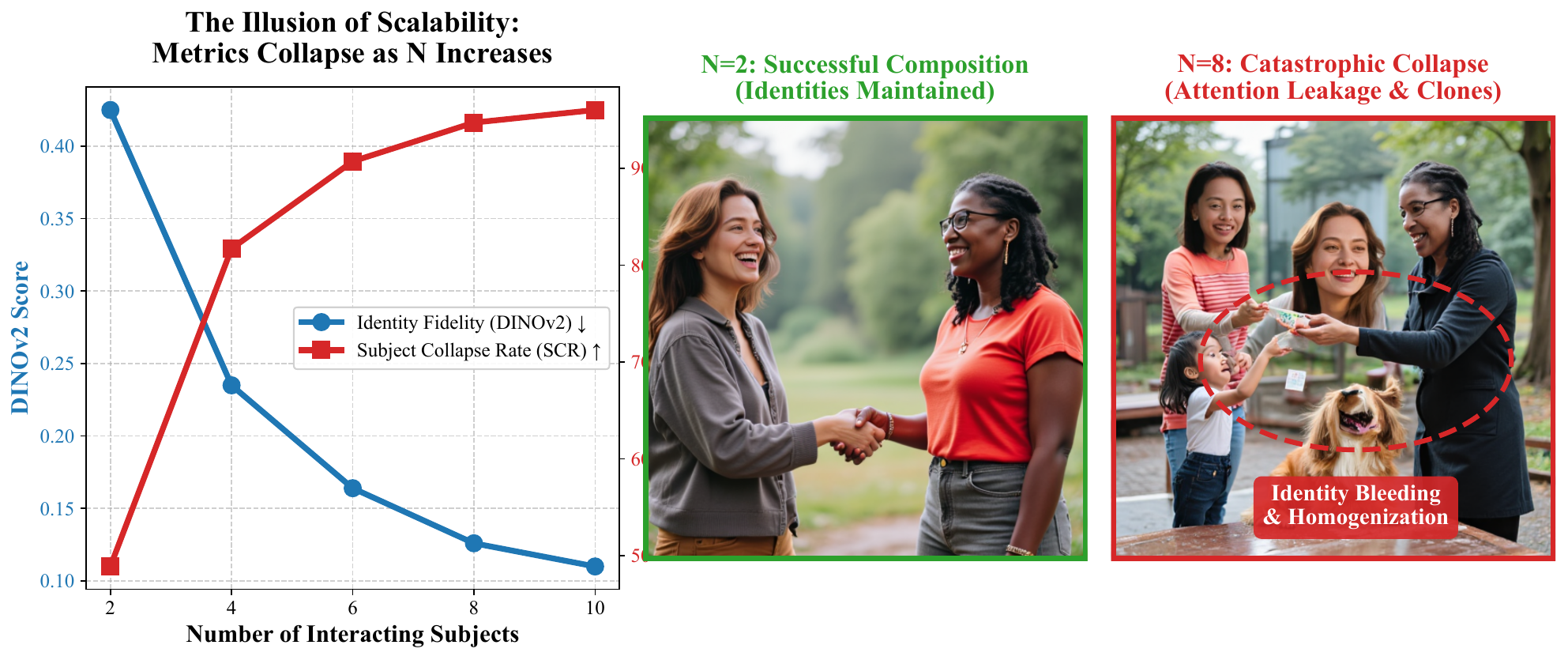}
  \caption{\textbf{Catastrophic Identity Collapse in Multi-Subject Personalization.} 
  (Left) Our quantitative analysis reveals the illusion of scalability: as the number of interacting subjects ($N$) increases, the fine-grained identity fidelity (DINOv2) plummets, and the Subject Collapse Rate (SCR) skyrockets to $>95\%$. 
  (Middle) At $N=2$, state-of-the-art models (e.g., MOSAIC~\cite{mosaic2024}) successfully maintain distinct identities and structural integrity. 
  (Right) When evaluated on our proposed stress-test benchmark at $N=8$, models experience severe attention leakage, resulting in identity bleeding and homogenization (generating clones of a single dominant identity).}
  \label{fig:teaser}
\end{figure*}

\section{Introduction}

Generative AI has fundamentally transformed visual content creation, with subject-driven text-to-image diffusion models demonstrating an unprecedented ability to generate customized images based on a few reference photos~\cite{ruiz2023dreambooth,kumari2023multi,gal2022image}. Early breakthroughs in personalization focused primarily on single-subject scenarios, successfully tackling the challenge of ``who appears in the scene.'' Recent advancements have further extended this capability to multi-subject composition, employing sophisticated techniques such as layout-to-region binding~\cite{contextgen2024,anyms2024} and non-invasive text-stream modulation to ensure that multiple referenced identities remain distinct~\cite{mosaic2024,xverse2024,psr2024}.

Despite these rapid advancements, the field remains constrained by a critical, yet underexplored, limitation: how well can these models compose many identities when they are deeply entangled through physical interaction and occlusion? In real-world applications, subjects rarely exist in isolated bounding boxes; they hug, grapple, and occlude one another. When multiple subjects interact, models based on architectures like DiT~\cite{peebles2023scalable} or FLUX~\cite{labs2025flux1kontextflowmatching} often suffer from severe \textit{attention leakage}. Because all tokens compete in a shared global attention space, identity cues, attributes, and structural priors from one subject frequently bleed into another. Consequently, what appears as a failure in identity preservation is, at its core, a failure in physically grounded attention routing.

Compounding this issue is the inadequacy of current evaluation metrics. The standard protocol for measuring personalization quality heavily relies on CLIP-based~\cite{radford2021learning} image-to-image similarity. However, CLIP is notoriously insensitive to fine-grained local features and structural integrity. In a multi-subject scene, even if a generated subject's face is completely distorted or incorrectly swapped with another identity, CLIP may still output an artificially high similarity score simply because the global semantic context matches. This metric blind spot masks the true severity of multi-subject entanglement and collapse, creating a false sense of progress.

To bridge this gap between semantic disentanglement and physical disentanglement, we present a systematic stress-test evaluation framework specifically designed for extreme multi-subject composition, directly addressing the core topics of benchmark and evaluation metrics for personalization quality. Our goal is to rigorously quantify the limits of current state-of-the-art models when faced with an increasing number of interacting identities. 

In summary, our main contributions are threefold:
\begin{itemize}
    \item \textbf{A Scalable Multi-Subject Benchmark}: We construct a comprehensive testing suite comprising prompts that scale from 2 to 10 interacting subjects, featuring various scene types including interaction, occlusion, and neutral layouts.
    \item \textbf{A Rigorous Evaluation Paradigm}: We shift the identity evaluation metric from the globally-biased CLIP to the structurally-sensitive DINOv2~\cite{oquab2023dinov2}. Furthermore, we propose the Subject Collapse Rate (SCR), a novel metric that explicitly quantifies the percentage of subjects that lose their identity in a generated scene.
    \item \textbf{Insightful Failure Analysis}: We evaluate three recent state-of-the-art multi-subject models (MOSAIC~\cite{mosaic2024}, XVerse~\cite{xverse2024}, PSR~\cite{psr2024}). Our findings reveal a catastrophic failure mode where SCR approaches 100\% as the subject count increases. We further uncover a critical trade-off: models actively sacrifice fine-grained identity fidelity (DINOv2) in favor of maintaining macro-level text alignment (CLIP-T), exposing fundamental limitations in current attention-routing architectures.
\end{itemize}

\section{Related Work}

\textbf{Foundations of Subject-Driven Generation.} 
The rapid evolution of text-to-image diffusion models~\cite{rombach2022high} has enabled profound new capabilities in personalized image generation. Foundational techniques like DreamBooth~\cite{ruiz2023dreambooth} and Textual Inversion~\cite{gal2022image} first demonstrated that a pre-trained model could learn novel subjects using only a few reference images. To improve efficiency and spatial control, subsequent works introduced parameter-efficient fine-tuning like LoRA~\cite{hu2021lora}, structural conditioning such as ControlNet~\cite{zhang2023adding}, and direct image-prompt adapters like IP-Adapter~\cite{ye2023ip}. While these methods achieve remarkable single-subject fidelity, they struggle with severe attribute binding and identity bleeding when tasked with generating multiple distinct subjects simultaneously.

\textbf{Multi-Subject Personalization.} 
Under DiT~\cite{peebles2023scalable} and FLUX~\cite{labs2025flux1kontextflowmatching} backbones, early multi-subject research primarily addressed a subject-centric question: how can multiple referenced identities remain distinct under shared generation dynamics? XVerse~\cite{xverse2024} is a representative method in this direction. It converts reference images into token-specific text-stream modulation offsets, injecting subject-specific control along the text-conditioning pathway rather than directly perturbing image latents. This non-invasive design improves identity controllability while avoiding direct corruption of the shared image representation. Similarly, MOSAIC~\cite{mosaic2024} tackles this problem from a representation-centric perspective. By enforcing semantic correspondence and feature disentanglement across references, it explicitly separates subjects in the feature space, reducing identity blending. Taken together, these methods move personalization beyond simple fidelity preservation toward explicit subject separation, establishing the first layer of controllable multi-subject generation.

\textbf{Layout-to-Region Binding.} 
Once subject identity is preserved, the next challenge is spatial grounding. This motivates a second line of work on layout-to-region binding, aiming to align references, prompts, and spatial regions. ContextGen~\cite{contextgen2024} introduces Contextual Layout Anchoring (CLA) and Identity Consistency Attention (ICA) to stabilize the correspondence between instances and target positions. AnyMS~\cite{anyms2024} addresses the same challenge in a training-free manner by decoupling attention at global and local levels to jointly maintain prompt alignment and layout fidelity. 3DIS-FLUX~\cite{3disflux2024} approaches the problem through depth-guided scene construction followed by FLUX-based rendering, utilizing depth as a compact structural prior. However, when the depth map underspecifies hidden geometry or front-back ordering, occlusion relations may still collapse. In essence, layout binding methods solve the ``where'' problem but only partially address the physically harder question of what should remain visible under overlap.

\textbf{Attention Leakage and Entanglement.} 
To understand multi-subject failures, it is crucial to examine the mechanics of attention leakage. In DiT-style generators, global attention is both a strength and a liability. While it enables rich long-range interaction, it also allows identity cues, attributes, colors, and poses from one subject to leak into another. This issue is especially pronounced in FLUX~\cite{labs2025flux1kontextflowmatching}, where the later single-stream blocks flatten text and image modalities into a tightly fused sequence, potentially losing the structural inductive bias for distinct instances. From this perspective, semantic entanglement is not merely an identity-preservation failure, but an \textit{attention-routing failure}. Existing methods attempt to mitigate this through various strategies, XVerse via text-stream modulation, MOSAIC via semantic disentanglement, and AnyMS via decoupled attention flows. This progression suggests that multi-subject generation is fundamentally constrained by how well the model can route attention without cross-instance contamination.

\textbf{Physical Interaction and Occlusion Reasoning.} 
The final and most challenging layer is physical interaction and occlusion reasoning. At this stage, the problem shifts to who occludes whom, whether partially hidden instances remain geometrically plausible, and whether the model respects front-back ordering under overlap. SeeThrough3D~\cite{seethrough3d2024} explicitly models which object should appear in front and how partial visibility should be rendered consistently, connecting naturally to the notion of amodal completion. LayerBind~\cite{chen2026layer} leverages the coarse-to-fine denoising dynamics of DiTs, establishing layout and visible order at early steps through layered initialization. Nevertheless, deeply entangled interactions, such as hugging, grappling, or interleaved articulated bodies, remain exceptionally difficult. The challenge lies not only in visibility ordering but also in fine-grained contact geometry, mutual deformation, and physically consistent completion under occlusion. Our work explicitly targets this gap, providing a rigorous benchmark and novel evaluation metrics to measure where current models fail in physically grounded interaction reasoning, pushing forward the frontier of multi-subject personalization.

\section{Multi-Subject Benchmark}

To systematically evaluate the limits of current subject-driven diffusion models in handling complex multi-entity compositions, we construct a rigorous, scalable stress-test benchmark. Unlike existing datasets that primarily focus on single or dual subjects in isolated settings, our benchmark is specifically designed to assess identity preservation and physical interaction reasoning as the number of subjects scales from 2 to 10. This directly addresses the critical need for dataset curation for benchmarking personalized generative models.

\subsection{Subject Pool Construction}
We establish a diverse and challenging subject pool by sourcing reference images from two well-established personalization datasets: the XVerse dataset~\cite{xverse2024} and the COSMISC dataset~\cite{mosaic2024}. These datasets provide a wide array of human and animal identities with varying attributes, poses, and clothing. We curate a unified subject pool by extracting high-quality reference images and assigning a unique identifier (e.g., S001, S002) to each distinct identity. This unified pool ensures that the models are tested on a consistent set of challenging visual features.

\subsection{Benchmark Design and Scalability}
\begin{figure}[t]
  \centering
  \includegraphics[width=\linewidth]{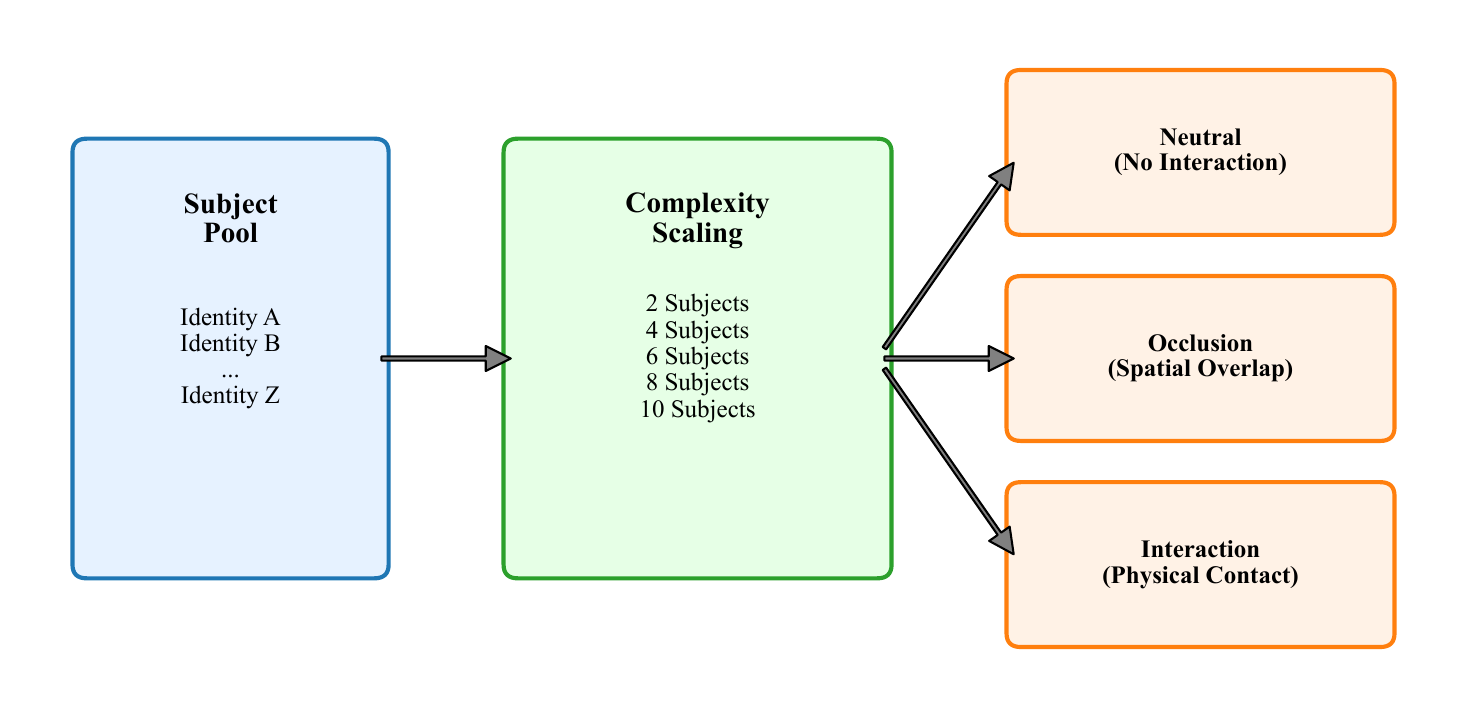}
  \caption{\textbf{Multi-Subject Benchmark Construction.} Our pipeline samples identities from a unified subject pool (left) to populate prompts of increasing complexity (2 to 10 subjects). Prompts are systematically categorized into Neutral, Occlusion, and Interaction scenarios (right) to isolate and test specific failure modes in attention routing and geometric reasoning.}
  \label{fig:benchmark_design}
\end{figure}

The core of our benchmark is a set of carefully crafted evaluation prompts designed to test both scalability and spatial reasoning. We define five distinct difficulty levels based on the subject count: \textbf{2, 4, 6, 8, and 10 subjects}, as illustrated in \textbf{Figure \ref{fig:benchmark_design}}.

For each subject level, we design 15 unique prompts, resulting in a total of 75 base prompts. To thoroughly investigate the impact of spatial entanglement on attention leakage, we strictly control the prompt IDs (1-75) and categorize them into three distinct scene types, ensuring an even distribution of 5 prompts per type within each subject level:
\begin{itemize}
    \item \textbf{Neutral (No Interaction)}: Subjects are placed in the same scene without physical overlap (e.g., \textit{``Subject A and Subject B standing next to each other''}). These prompts test the model's baseline ability to prevent identity bleeding across spatially separated regions.
    \item \textbf{Occlusion}: One subject partially blocks the view of another (e.g., \textit{``Subject A standing behind Subject B''}). These prompts test the model's amodal completion and depth ordering capabilities.
    \item \textbf{Interaction}: Subjects are physically engaged with one another (e.g., \textit{``Subject A hugging Subject B''} or \textit{``Subject A hugging Subject B while Subject C stands behind Subject D''}). These are the most challenging prompts, testing the model's ability to maintain fine-grained contact geometry and identity under severe attention entanglement.
\end{itemize}

\subsection{Evaluation Protocol}
We evaluate three recent state-of-the-art multi-subject personalization models: \textbf{XVerse}~\cite{xverse2024}, \textbf{MOSAIC}~\cite{mosaic2024}, and \textbf{PSR}~\cite{psr2024}. These models were specifically selected because they represent the frontier of non-invasive, attention-modulation techniques designed to overcome the limitations of naive region-masking, making them the strongest candidates for complex physical interactions. For each of the 75 prompts, the models take the text prompt and the corresponding subject reference images as input. To account for the stochastic nature of diffusion models, we generate images using 3 random seeds per prompt. This yields a total of 225 generated images per model (675 images across all three models), providing a statistically robust foundation for our quantitative analysis. All generation metadata, including prompt IDs, subject assignments, and output paths, are strictly tracked to ensure reproducible evaluation.

\section{Evaluation Metrics}

A central contribution of our benchmark is identifying the limitations of standard personalization metrics when applied to multi-subject composition. Existing protocols heavily rely on global CLIP embeddings~\cite{radford2021learning}. However, global embeddings are highly tolerant of local structural distortions and identity bleeding, rendering them inadequate for measuring severe identity entanglement. To provide a rigorous and comprehensive assessment that aligns with the need for advanced evaluation metrics for personalization quality, we introduce a multi-tiered evaluation framework.

\subsection{Text-Image Alignment (CLIP-T)}
To measure how well the generated image aligns with the semantic intent of the text prompt, we compute the cosine similarity between the CLIP~\cite{radford2021learning} text embedding of the prompt and the CLIP image embedding of the generated output. 

While CLIP-T is a standard metric for semantic fidelity, we observe a critical caveat in multi-subject scenarios: models may achieve high CLIP-T scores by generating a generic group of people that matches the macro-semantics of the prompt (e.g., ``a group of people''), while completely failing to preserve the specific identities requested. Therefore, CLIP-T must be analyzed in conjunction with fine-grained identity metrics.

\subsection{Identity Preservation: From CLIP to DINOv2}
Traditionally, identity preservation is measured by computing the cosine similarity between the CLIP image embedding of the generated subject and the reference image (denoted as \textbf{CLIP-I}). However, our experiments reveal that CLIP-I scores remain artificially high even when subjects undergo severe identity bleeding or facial distortion.

To capture these local structural failures, we shift our primary identity evaluation from CLIP to \textbf{DINOv2}~\cite{oquab2023dinov2}. Unlike CLIP, which is trained primarily via contrastive language-image pre-training and tends to prioritize global semantic layout (e.g., ``a person with glasses''), DINOv2 is a self-supervised vision transformer trained via image-level and patch-level objectives. This training paradigm grants DINOv2 an exceptional sensitivity to fine-grained local features, part-level correspondence, and structural geometry, making it significantly more robust for evaluating identity preservation under complex physical interactions where semantic features might otherwise bleed. While earlier versions like DINOv1~\cite{caron2021emerging} also capture structural priors, DINOv2 leverages a significantly larger and curated pre-training dataset along with an improved objective function, resulting in more discriminative and stable feature embeddings for complex multi-entity scenes.

While a rigorous subject-level evaluation would ideally require instance segmentation masks to isolate each generated subject, such masks are notoriously difficult to obtain accurately in highly entangled multi-subject scenes (e.g., severe occlusion or physical interaction). Therefore, as an established proxy for overall identity fidelity, we compute the DINOv2 image embedding of the \textit{entire} generated image $I_{gen}$ and calculate its average cosine similarity against the embeddings of all $N$ individual reference images $\{I_{ref}^{(1)}, I_{ref}^{(2)}, \dots, I_{ref}^{(N)}\}$:

\begin{equation}
\text{DINOv2 Score} = \frac{1}{N} \sum_{i=1}^{N} \cos(\text{DINOv2}(I_{gen}), \text{DINOv2}(I_{ref}^{(i)}))
\end{equation}

Although comparing a multi-subject scene embedding against single-subject reference embeddings introduces scene complexity into the score, this formulation serves as a highly effective penalty mechanism. It strictly demands that the structural identity of \textit{every} requested subject be strongly represented in the global feature space.

\subsection{Subject Collapse Rate (SCR)}
Average similarity scores can obscure catastrophic failures of individual subjects within a complex scene. For instance, in an 8-subject image, if 7 subjects are perfectly generated but 1 subject is completely missing or morphed into another identity, the mean DINOv2 score might still appear acceptable.

\begin{figure}[t]
  \centering
  \includegraphics[width=\linewidth]{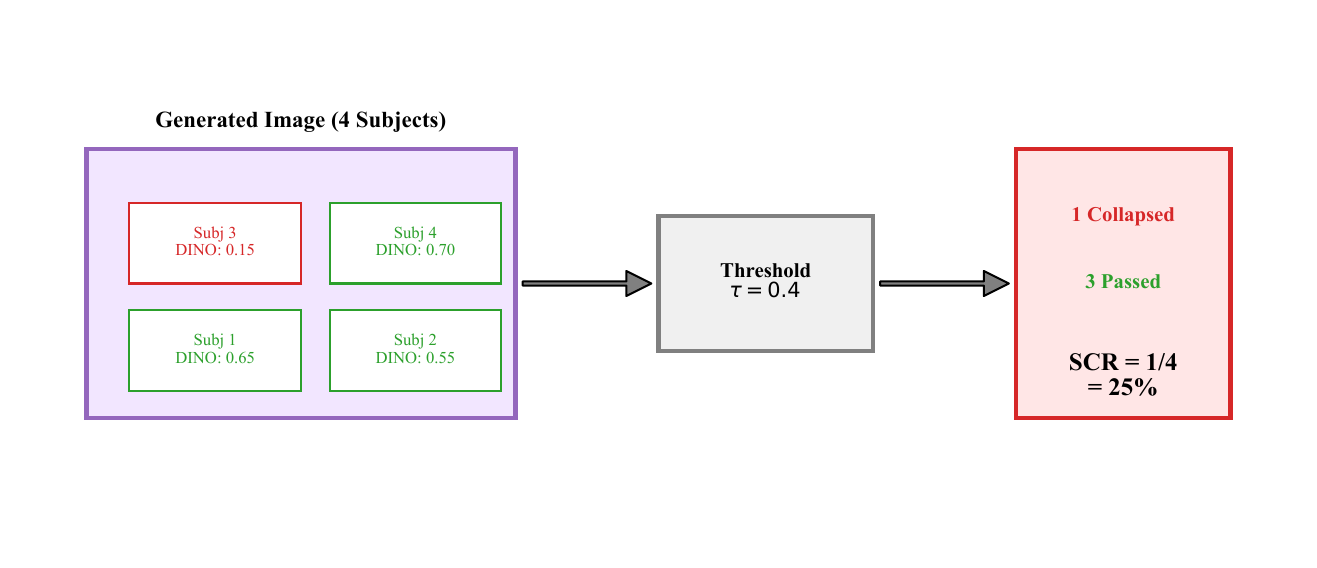}
  \caption{\textbf{Subject Collapse Rate (SCR).} Unlike average similarity scores which mask individual failures, SCR explicitly counts the proportion of subjects whose DINOv2 identity similarity falls below a strict threshold $\tau$. This provides a more realistic measure of multi-subject entanglement.}
  \label{fig:scr_illustration}
\end{figure}

To explicitly quantify these localized failures, we propose the \textbf{Subject Collapse Rate (SCR)}, conceptually illustrated in \textbf{Figure \ref{fig:scr_illustration}}. While this metric still utilizes the scene-level generated embedding, it shifts the evaluation from a continuous average to a strict discrete thresholding. We define a subject as "collapsed" if its DINOv2 cosine similarity with the reference image falls below a predefined threshold $\tau$. The SCR for a given generated image is defined as the ratio of collapsed subjects to the total number of subjects:

\begin{equation}
\text{SCR}_{@\tau} = \frac{1}{N} \sum_{i=1}^{N} \mathds{1}\Big[\cos(\text{DINOv2}(I_{gen}), \text{DINOv2}(I_{ref}^{(i)})) < \tau\Big]
\end{equation}

where $\mathds{1}[\cdot]$ is the indicator function. Because DINOv2 similarities typically occupy a lower and more discriminative numerical range than CLIP, we employ strict thresholds $\tau \in \{0.4, 0.5, 0.6\}$, which empirical visual inspection confirms align with severe human-perceivable identity loss. Importantly, this metric naturally penalizes the ``Homogenization'' failure mode: if a model generates multiple clones of a single dominant subject, only that subject's reference will yield a high similarity, while the remaining $N-1$ subjects will fall below the threshold, correctly driving the SCR towards 1.0. A lower SCR indicates better multi-subject preservation, while an SCR approaching 1.0 signifies a complete collapse of personalization.

\section{Experiments}

In this section, we present the quantitative and qualitative results of our evaluation. We benchmark three state-of-the-art multi-subject personalization models: \textbf{MOSAIC}~\cite{mosaic2024}, \textbf{XVerse}~\cite{xverse2024}, and \textbf{PSR}~\cite{psr2024}, across five difficulty levels ranging from 2 to 10 interacting subjects.

\subsection{Quantitative Results: The Limits of Personalization}

\begin{figure*}[t]
  \centering
  \includegraphics[width=\textwidth]{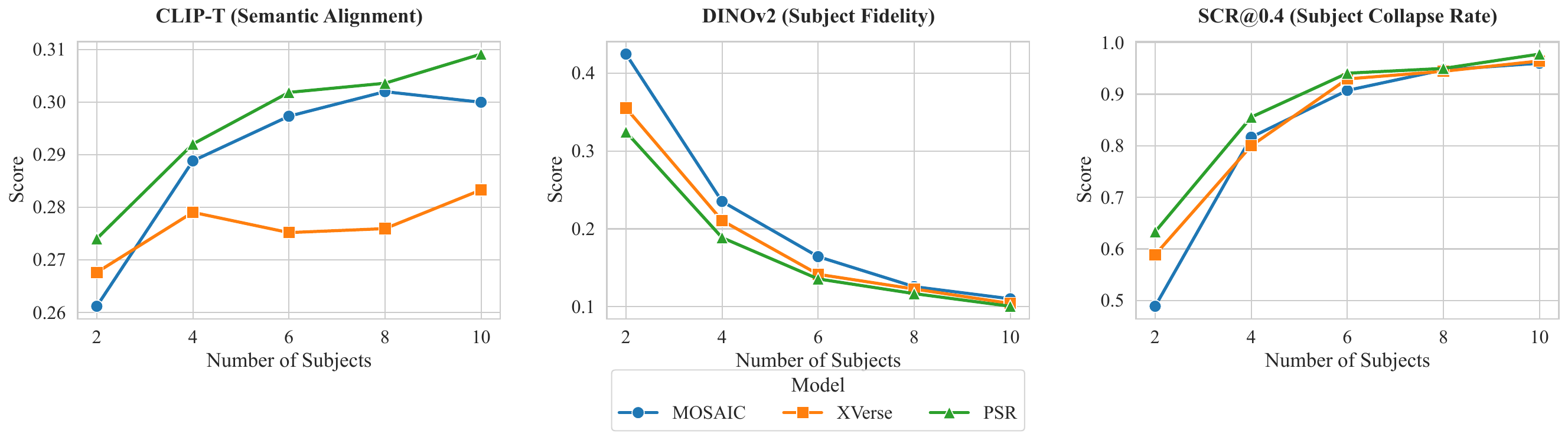}
  \caption{\textbf{Performance Trends across Subject Counts.} (Left) DINOv2 identity similarity exhibits a precipitous drop for all models as scenes become denser. (Right) Subject Collapse Rate (SCR@0.4) skyrockets from $\sim$50\% at 2 subjects to nearly 100\% at 8-10 subjects, highlighting a fundamental scalability bottleneck.}
  \label{fig:trend_charts}
\end{figure*}

To rigorously assess model performance, we first examine the overall trend of identity preservation as the subject count increases. The quantitative results, averaged across all scene types, are summarized in \textbf{Table \ref{tab:main_results}} and visualized in \textbf{Figure \ref{fig:trend_charts}}.

\begin{table*}[t]
\centering
\caption{\textbf{Quantitative Evaluation on Multi-Subject Personalization Benchmark.} We report CLIP-T, CLIP-I, DINOv2, and Subject Collapse Rate (SCR@0.4) across varying subject counts (2 to 10). As subject count increases, DINOv2 sharply declines and SCR approaches 100\%, indicating catastrophic identity failure.}
\label{tab:main_results}
\resizebox{\textwidth}{!}{
\begin{tabular}{l|cccc|cccc|cccc}
\toprule
\multirow{2}{*}{\textbf{Subjects}} & \multicolumn{4}{c|}{\textbf{MOSAIC}~\cite{mosaic2024}} & \multicolumn{4}{c|}{\textbf{XVerse}~\cite{xverse2024}} & \multicolumn{4}{c}{\textbf{PSR}~\cite{psr2024}} \\
\cmidrule(lr){2-5} \cmidrule(lr){6-9} \cmidrule(lr){10-13}
 & CLIP-T $\uparrow$ & CLIP-I $\uparrow$ & DINOv2 $\uparrow$ & SCR $\downarrow$ & CLIP-T $\uparrow$ & CLIP-I $\uparrow$ & DINOv2 $\uparrow$ & SCR $\downarrow$ & CLIP-T $\uparrow$ & CLIP-I $\uparrow$ & DINOv2 $\uparrow$ & SCR $\downarrow$ \\
\midrule
2 & 0.261 & 0.695 & 0.425 & 48.9\% & 0.268 & 0.646 & 0.355 & 58.9\% & 0.274 & 0.647 & 0.325 & 63.3\% \\
4 & 0.289 & 0.586 & 0.235 & 81.7\% & 0.279 & 0.593 & 0.211 & 80.0\% & 0.292 & 0.568 & 0.189 & 85.6\% \\
6 & 0.297 & 0.535 & 0.164 & 90.7\% & 0.275 & 0.550 & 0.142 & 93.0\% & 0.302 & 0.537 & 0.136 & 94.1\% \\
8 & 0.302 & 0.517 & 0.126 & 94.7\% & 0.276 & 0.532 & 0.123 & 94.4\% & 0.304 & 0.517 & 0.117 & 95.0\% \\
10 & 0.300 & 0.504 & 0.110 & 96.0\% & 0.283 & 0.524 & 0.104 & 96.4\% & 0.309 & 0.500 & 0.101 & 97.8\% \\
\bottomrule
\end{tabular}
}
\end{table*}

\textbf{Catastrophic Identity Forgetting.} 
The most striking finding from our benchmark is the catastrophic collapse of identity fidelity when scaling beyond 4 subjects. As shown in the DINOv2 scores, all models experience a precipitous drop. For instance, MOSAIC~\cite{mosaic2024}, which performs best at the 2-subject level with a DINOv2 score of 0.425, plummets to 0.110 at 10 subjects. XVerse~\cite{xverse2024} and PSR~\cite{psr2024} follow a nearly identical trajectory, dropping from 0.355 and 0.325 to 0.104 and 0.101, respectively. 

This failure is further magnified when examining the Subject Collapse Rate (SCR). At the most lenient threshold ($\tau=0.4$), MOSAIC exhibits a 48.9\% collapse rate even with just 2 subjects. When pushed to 8 subjects, the SCR skyrockets to 94.7\% for MOSAIC, 94.4\% for XVerse, and 95.0\% for PSR. At 10 subjects, the SCR for all models exceeds 96\%, indicating that nearly every generated subject has lost its intended identity. This quantitative evidence strongly supports our hypothesis: current attention-routing mechanisms are fundamentally inadequate for resolving dense physical entanglement.

\begin{figure}[t]
  \centering
  \includegraphics[width=0.85\linewidth]{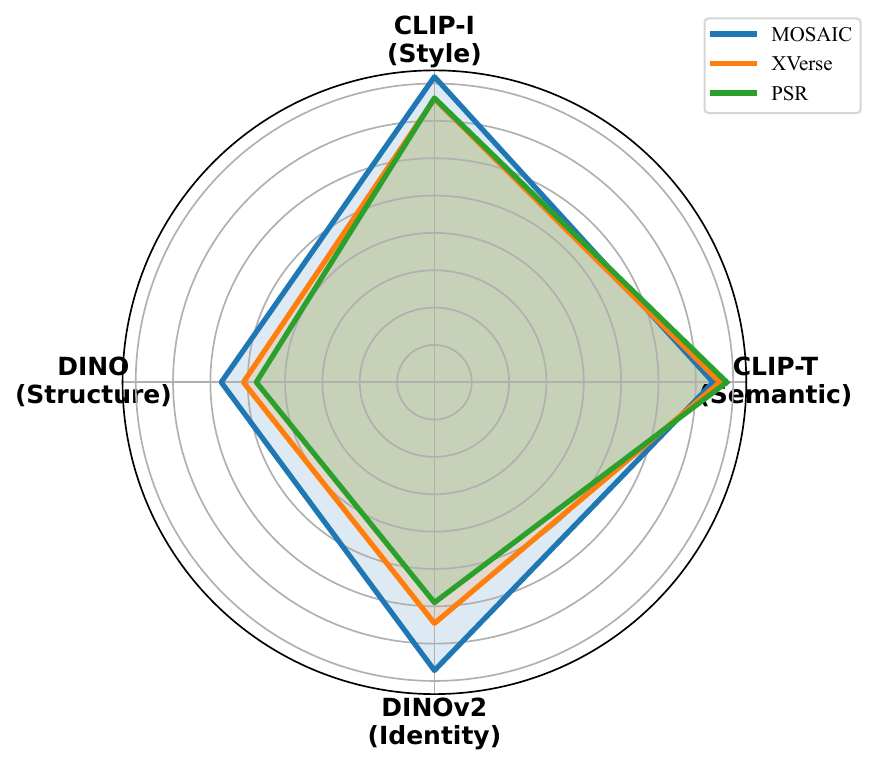}
  \caption{\textbf{Comprehensive Metric Radar (2 Subjects).} A multi-dimensional comparison of MOSAIC, XVerse, and PSR. While all models maintain high semantic alignment (CLIP-T), MOSAIC shows a distinct advantage in structural (DINO) and fine-grained identity (DINOv2) preservation.}
  \label{fig:radar_plot}
\end{figure}

\textbf{Model Comparison.} 
While all models fail at extreme subject counts, \textbf{MOSAIC} demonstrates a noticeably stronger baseline in low-complexity scenarios. At 2 subjects, MOSAIC achieves the highest DINOv2 score (0.425) and the lowest SCR (48.9\%), outperforming XVerse (0.355 / 58.9\%) and PSR (0.325 / 63.3\%). This multifaceted performance profile is further detailed in \textbf{Figure \ref{fig:radar_plot}}, which compares the semantic, style, structure, and identity metrics simultaneously. This suggests that MOSAIC's representation-centric disentanglement provides a more robust defense against attention leakage when the scene is relatively sparse. However, this advantage quickly dissipates as the subject count reaches 6 and beyond, underscoring the universal scalability bottleneck in current DiT/FLUX architectures~\cite{peebles2023scalable,labs2025flux1kontextflowmatching}.

\subsection{The Trade-off: Semantic vs. Physical Disentanglement}

A counter-intuitive phenomenon emerges when we analyze the Text-Image Alignment (CLIP-T)~\cite{radford2021learning} scores alongside the identity metrics. While DINOv2~\cite{oquab2023dinov2} scores plummet as the subject count increases, the CLIP-T scores for all models actually exhibit a slight upward trend. For example, PSR's CLIP-T score rises from 0.274 (2 subjects) to 0.309 (10 subjects), and MOSAIC rises from 0.261 to 0.300.

This exposes a critical trade-off mechanism inherent in current diffusion models. When faced with the overwhelming complexity of generating 8 or 10 distinct, interacting individuals, the models actively ``take a shortcut.'' Instead of attempting to faithfully reconstruct each specific identity, which would require precise, uncorrupted attention routing, the models revert to generating a generic ``group of people'' that satisfies the macro-level semantic constraints of the prompt. Consequently, the global CLIP-T score improves, but at the total expense of personalized identity.

\subsection{Qualitative Failure Analysis}

To better understand \textit{how} these models fail, we perform a qualitative visual analysis of the generated scenes.

\begin{figure*}[t]
  \centering
  \includegraphics[width=\textwidth]{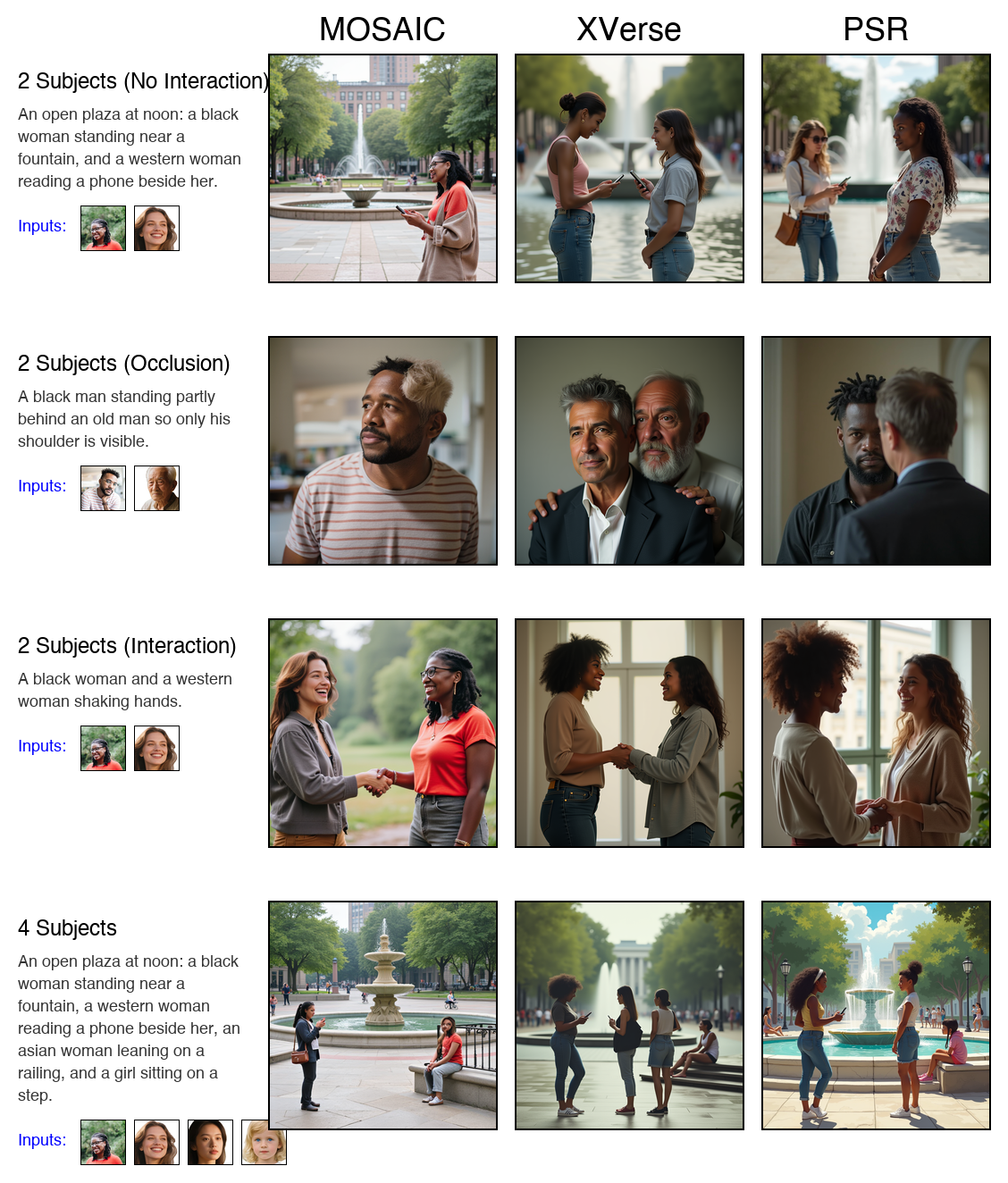}
  \caption{\textbf{Qualitative Failure Analysis.} Comparison between different models across various subject counts and scene types. In denser scenes (4+ subjects) or complex interactions, models exhibit severe (1) Identity Bleeding, where features merge; (2) Homogenization, generating clones of a single dominant identity; and (3) Amodal Collapse, resulting in malformed geometry under occlusion.}
  \label{fig:qualitative_failures}
\end{figure*}

\begin{figure*}[t]
  \centering
  \includegraphics[width=\textwidth]{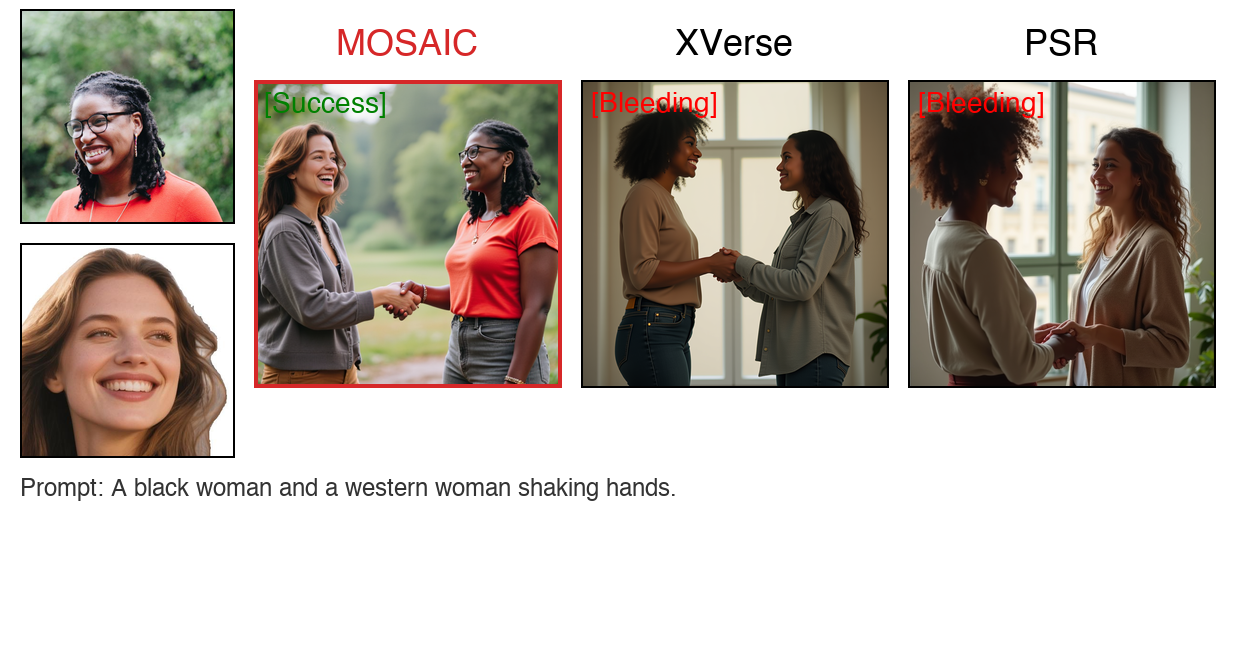}
  \caption{\textbf{Detailed Case Analysis of Identity Bleeding.} (Left) Reference images of two distinct subjects. (Right) Generation results for a complex interaction prompt (``A black woman and a western woman shaking hands''). While the MOSAIC baseline successfully maintains identity boundaries (marked with green check), XVerse and PSR suffer from severe identity bleeding, where the features of Subject A contaminate the spatial region of Subject B (marked with red cross).}
  \label{fig:case_analysis}
\end{figure*}

Visual inspection corroborates our quantitative findings and reveals three primary failure modes during multi-subject interaction:
\begin{enumerate}
    \item \textbf{Identity Bleeding}: Features from one subject (e.g., clothing color, facial structure, glasses) seamlessly bleed into an adjacent subject, especially during physical contact (e.g., hugging).
    \item \textbf{Homogenization}: In dense scenes (8-10 subjects), the model often generates multiple copies of the \textit{same} dominant reference identity, completely ignoring the other requested subjects.
    \item \textbf{Amodal Collapse}: When subjects are occluded, the model frequently fails to render the hidden geometry correctly, resulting in fused limbs or disembodied floating heads.
\end{enumerate}

\section{Discussion}

Our comprehensive evaluation uncovers several fundamental bottlenecks in current multi-subject personalization models, pointing to critical areas for future research within the generative AI community.

\subsection{The ``Semantic Shortcut'' Hypothesis}
The divergent trends between CLIP-T~\cite{radford2021learning} (which slightly improves) and DINOv2~\cite{oquab2023dinov2} (which drastically drops) as subject count increases suggest that models are not merely ``failing'' to generate images, but are actively optimizing for the wrong objective. In DiT~\cite{peebles2023scalable} and FLUX~\cite{labs2025flux1kontextflowmatching} architectures, when the global attention mechanism is overwhelmed by multiple, competing identity embeddings, the model defaults to the strongest available prior: the global text prompt. We term this the ``semantic shortcut.'' The model synthesizes a structurally plausible scene of ``many people'' to satisfy the text encoder, completely washing out the localized, high-frequency details required for identity preservation. This highlights a dangerous flaw in relying solely on CLIP for generative evaluation.

\subsection{From Semantic to Physical Disentanglement}
While recent methods like MOSAIC~\cite{mosaic2024} and XVerse~\cite{xverse2024} have made significant strides in semantic disentanglement (e.g., ensuring ``dog'' features don't mix with ``cat'' features in latent space), our benchmark reveals that they still fail at \textit{physical disentanglement}. When subjects interact (e.g., hugging) or occlude one another, the boundary between their spatial regions becomes blurred. Because current models lack explicit 3D reasoning or amodal completion mechanisms~\cite{seethrough3d2024}, the physical overlap directly translates into attention leakage in the 2D feature maps. Solving ``who is who'' is insufficient if the model cannot resolve ``who is in front of whom.'' Future directions could involve incorporating direct geometric priors into transformer backbones, as demonstrated by recent breakthroughs in feed-forward 3D networks like VGGT~\cite{wang2025vggt}, temporal-spatial consistency models like NWM~\cite{bar2025navigation}, and highly expressive rendering primitives such as Student Splatting and Scooping~\cite{zhu20253d}, to enforce physical boundaries and temporal-spatial consistency during the denoising process.

\subsection{Limitations}
We acknowledge certain limitations in our current benchmark. First, our evaluation is primarily focused on human and animal subjects; evaluating multi-object compositions involving inanimate objects with rigid geometries may yield different failure modes. Second, our scene types (neutral, occlusion, interaction) are defined via prompt engineering rather than explicit 3D layout controls~\cite{contextgen2024,anyms2024}, meaning the severity of occlusion is left to the model's interpretation.

\section{Conclusion}

In this paper, we presented a rigorous stress-test benchmark to evaluate the limits of subject-driven diffusion models in multi-entity compositions, directly addressing the critical need for robust evaluation metrics in the personalization domain. We demonstrated that standard CLIP-based metrics are dangerously blind to local identity collapse, and we proposed the Subject Collapse Rate (SCR) based on DINOv2 to quantify this failure. Our extensive evaluation of SOTA models (MOSAIC~\cite{mosaic2024}, XVerse~\cite{xverse2024}, PSR~\cite{psr2024}) yielded a sobering conclusion: while models can handle 2-4 isolated subjects, they suffer from catastrophic identity forgetting, with SCR approaching 100\%, when forced to compose 6 to 10 interacting identities. This failure exposes a fundamental flaw in current global attention routing mechanisms, which actively sacrifice localized physical fidelity to satisfy macro-level semantic alignment. To overcome this bottleneck, future architectures must move beyond global token modulation and explore explicit representation disentanglement or part-level correspondence mechanisms. We hope our benchmark and the proposed SCR metric will serve as a clear, demanding target for the next generation of physically-grounded personalization models.

{
    \small
    \bibliographystyle{ieeenat_fullname}
    \bibliography{main}
}


\end{document}